\documentclass[twocolumn,american]{revtex4-1}
\usepackage[latin9]{inputenc}
\usepackage[letterpaper]{geometry}
\usepackage{color}
\usepackage{babel}
\usepackage{units}
\usepackage{amsmath}
\usepackage{graphicx}
\usepackage[unicode=true]
 {hyperref}

\makeatletter
 \@ifundefined{textcolor}{}
 {%
   \definecolor{BLACK}{gray}{0}
   \definecolor{WHITE}{gray}{1}
   \definecolor{RED}{rgb}{1,0,0}
   \definecolor{GREEN}{rgb}{0,1,0}
   \definecolor{BLUE}{rgb}{0,0,1}
   \definecolor{CYAN}{cmyk}{1,0,0,0}
   \definecolor{MAGENTA}{cmyk}{0,1,0,0}
   \definecolor{YELLOW}{cmyk}{0,0,1,0}
 }

\usepackage{placeins}
\usepackage{units}
\usepackage[T1]{fontenc} 
\usepackage{amsmath}
\usepackage{mathtools}
\allowdisplaybreaks %does the page break in align environments
\usepackage{amssymb}
\usepackage{graphicx}
\usepackage[bf, small, centerlast, font=footnotesize]{caption}
\usepackage{wrapfig}
\usepackage{footnote}
\usepackage{listings}
\usepackage{color}
\usepackage[wide]{sidecap}
\usepackage{epstopdf}
\usepackage{marvosym}
\usepackage{hyperref}
\usepackage{bbm}

\makeatother

\begin{document}

\title{VAMPnets for deep learning of molecular kinetics}

\author{Andreas Mardt$^{1}$, Luca Pasquali$^{1}$, Hao Wu and Frank No\'{e}$^{*}$}

\affiliation{Freie Universit\"{a}t Berlin, Department of Mathematics and Computer
Science, \\
Arnimallee 6, 14195 Berlin, Germany\\
$1$ equal contribution\hspace{0.3cm} $*$ correspondence to frank.noe@fu-berlin.de}
\begin{abstract}
There is an increasing demand for computing the relevant structures,
equilibria and long-timescale kinetics of biomolecular processes,
such as protein-drug binding, from high-throughput molecular dynamics
simulations. Current methods employ transformation of simulated coordinates
into structural features, dimension reduction, clustering the dimension-reduced
data, and estimation of a Markov state model or related model of the
interconversion rates between molecular structures. This handcrafted
approach demands a substantial amount of modeling expertise, as poor
decisions at any step will lead to large modeling errors. Here we
employ the variational approach for Markov processes (VAMP) to develop
a deep learning framework for molecular kinetics using neural networks,
dubbed VAMPnets. A VAMPnet encodes the entire mapping from molecular
coordinates to Markov states, thus combining the whole data processing
pipeline in a single end-to-end framework. Our method performs equally
or better than state-of-the art Markov modeling methods and provides
easily interpretable few-state kinetic models.
\end{abstract}
\maketitle

\section*{Introduction}

\newcounter{supplfig}
\refstepcounter{supplfig}\label{fig:1dfolding}
\refstepcounter{supplfig}\label{fig:tau_conv}
\refstepcounter{supplfig}\label{fig:small_ala}

The rapid advances in computing power and simulation technologies
for molecular dynamics (MD) of biomolecules and fluids \cite{LindorffLarsenEtAl_Science11_AntonFolding,PlattnerEtAl_NatChem17_BarBar,KohlhoffEtAl_NatChem14_GPCR-MSM,DoerrEtAl_JCTC16_HTMD},
and \emph{ab initio} MD of small molecules and materials \cite{UfimtsevMartinez_CSE08_Terachem,MarxHutter_NICseries00_AbInitioMD},
allow the generation of extensive simulation data of complex molecular
systems. Thus, it is of high interest to automatically extract statistically
relevant information, including stationary, kinetic and mechanistic
properties.\\
The Markov modeling approach \cite{SchuetteFischerHuisingaDeuflhard_JCompPhys151_146,PrinzEtAl_JCP10_MSM1,SwopePiteraSuits_JPCB108_6571,NoeHorenkeSchutteSmith_JCP07_Metastability,ChoderaEtAl_JCP07,BucheteHummer_JPCB08}
has been a driving force in the development of kinetic modeling techniques
from MD mass data, chiefly as it facilitates a divide-and-conquer
approach to integrate short, distributed MD simulations into a model
of the long-timescale behavior. State-of-the-art analysis approaches
and software packages \cite{SchererEtAl_JCTC15_EMMA2,DoerrEtAl_JCTC16_HTMD,HarriganEtAl_BJ17_MSMbuilder}
operate by a sequence, or pipeline, of multiple processing steps,
that has been engineered by practitioners over the last decade. The
first step of a typical processing pipeline is featurization, where
the MD coordinates are either aligned (removing translation and rotation
of the molecule of interest) or transformed into internal coordinates
such as residue distances, contact maps or torsion angles \cite{HumphreyDalkeSchulten_JMG96_VMD,McGibbon_BJ15_MDTraj,SchererEtAl_JCTC15_EMMA2,DoerrEtAl_JCTC16_HTMD}.
This is followed by a dimension\textbf{ }reduction, in which the dimension
is reduced to much fewer (typically 2-100) slow collective variables
(CVs), often based on the variational approach or conformation dynamics
\cite{NoeNueske_MMS13_VariationalApproach,NueskeEtAl_JCTC14_Variational},
time-lagged independent component analysis (TICA) \cite{PerezEtAl_JCP13_TICA,SchwantesPande_JCTC13_TICA},
blind source separation \cite{Molgedey_94,ZieheMueller_ICANN98_TDSEP,HarmelingEtAl_NeurComput03_KernelTDSEP}
or dynamic mode decomposition \cite{Mezic_NonlinDyn05_Koopman,SchmidSesterhenn_APS08_DMD,TuEtAl_JCD14_ExactDMD,WilliamsKevrekidisRowley_JNS15_EDMD,WuEtAl_JCP17_VariationalKoopman}
\textendash{} see \cite{NoeClementi_COSB17_SlowCVs,KlusEtAl_JNS17_DataDriven}
for an overview. The resulting coordinates may be scaled, in order
to embed them in a metric space whose distances correspond to some
form of dynamical distance \cite{NoeClementi_JCTC15_KineticMap,NoeClementi_JCTC16_KineticMap2}.
The resulting metric space is discretized by clustering the projected
data using hard or fuzzy data-based clustering methods \cite{BowmanPandeNoe_MSMBook,SchererEtAl_JCTC15_EMMA2,HusicPande_JCTC17_Ward,SheongEtAl_JCTC15_APM,ChoderaEtAl_JCP07,WuNoe_JCP15_GMTM,WeberFackeldeySchuette_JCP17_SetfreeMSM},
typically resulting in 100-1000 discrete states. A transition matrix
or rate matrix describing the transition probabilities or rate between
the discrete states at some lag time $\tau$ is then estimated \cite{Bowman_JCP09_Villin,PrinzEtAl_JCP10_MSM1,BucheteHummer_JPCB08,TrendelkampSchroerEtAl_InPrep_revMSM}
(alternatively, a Koopman model can be built after the dimension reduction
\cite{WilliamsKevrekidisRowley_JNS15_EDMD,WuEtAl_JCP17_VariationalKoopman}).
The final step towards an easily interpretable kinetic model is coarse-graining
of the estimated Markov state model (MSM) down to a few states \cite{KubeWeber_JCP07_CoarseGraining,YaoHuang_JCP13_Nystrom,FackeldeyWeber_WIAS17_GenPCCA,GerberHorenko_PNAS17_Categorial,HummerSzabo_JPCB15_CoarseGraining,OrioliFaccioli_JCP16_CoarseMSM,NoeEtAl_PMMHMM_JCP13}.

This sequence of analysis steps has been developed by combining physico-chemical
intuition and technical experience gathered in the last $\sim$10
years. Although each of the steps in the above pipeline appears meaningful,
there is no fundamental reason why this or any other given analysis
pipeline should be optimal. More dramatically, the success of kinetic
modeling currently relies on substantial technical expertise of the
modeler, as suboptimal decisions in each step may deteriorate the
result. As an example, failure to select suitable features in step
1 will almost certainly lead to large modeling errors. 

An important step towards selecting optimal models (parameters) and
modeling procedures (hyper-parameters) has been the development of
the variational approach for conformation dynamics (VAC) \cite{NoeNueske_MMS13_VariationalApproach,NueskeEtAl_JCTC14_Variational},
which offers a way to define scores that measure the optimality of
a given kinetic model compared to the (unknown) MD operator that governs
the true kinetics underlying the data. The VAC has recently been generalized
to the variational approach for Markov processes (VAMP), which allows
to optimize models of arbitrary Markov processes, including nonreversible
and non-stationary dynamics \cite{WuNoe_VAMP}. The VAC has been employed
using cross-validation in order to make optimal hyper-parameter choices
within the analysis pipeline described above while avoiding overfitting
\cite{McGibbonPande_JCP15_CrossValidation,HusicPande_JCTC17_Ward}.
However, a variational score is not only useful to optimize the steps
of a given analysis pipeline, but in fact allows us to replace the
entire pipeline with a more general learning structure.

Here we develop a deep learning structure that is in principle able
to replace the entire analysis pipeline above. Deep learning has been
very successful in a broad range of data analysis and learning problems
\cite{lecun2015deep,KrizhevskySutskeverHinton_NIPS12_ImageNetCNN,MnihEtAl_Nature15_DeepReinforcement}.
A feedforward deep neural network is a structure that can learn a
complex, nonlinear function $\mathbf{y}=F(\mathbf{x})$. In order
to train the network a scoring or loss function is needed that is
maximized or minimized, respectively. Here we develop \emph{VAMPnets},
a neural network architecture that can be trained by maximizing a
VAMP variational score. VAMPnets contain two network lobes that transform
the molecular configurations found at a time delay $\tau$ along the
simulation trajectories. Compared to previous attempts to include
``depth'' or ``hierarchy'' into the analysis method \cite{PerezNoe_JCTC16_hTICA,NueskeEtAl_JCP15_Tensor},
VAMPnets combine the tasks of featurization, dimension reduction,
discretization and coarse-grained kinetic modeling into a single end-to-end
learning framework. We demonstrate the performance of our networks
using a variety of stochastic models and datasets, including a protein
folding dataset. The results are competitive with and sometimes surpass
the state-of-the-art handcrafted analysis pipeline. Given the rapid
improvements of training efficiency and accuracy of deep neural networks
seen in a broad range of disciplines, it is likely that follow-up
works can lead to superior kinetic models.

\section*{Results}

\subsection{Variational principle for Markov processes}

\label{subsec:VAMP-principle}MD can be theoretically described as
a Markov process $\{\mathbf{x}_{t}\}$ in the full state space $\Omega$.
For a given potential energy function, the simulation setup (e.g.
periodic boundaries) and the time-step integrator used, the dynamics
are fully characterized by a transition density $p_{\tau}\left(\mathbf{x},\mathbf{y}\right)$,
i.e. the probability density that a MD trajectory will be found at
configuration $\mathbf{y}$ given that it was at configuration $\mathbf{x}$
a time lag $\tau$ before. Markovianity implies that the $\mathbf{y}$
can be sampled by knowing $\mathbf{x}$ alone, without the knowledge
of previous time-steps. While the dynamics might be highly nonlinear
in the variables $\mathbf{x}_{t}$, Koopman theory \cite{Koopman_PNAS31_Koopman,Mezic_NonlinDyn05_Koopman}
tells us that there is a transformation of the original variables
into some features or latent variables that, on average, evolve according
to a linear transformation. In mathematical terms, there exist transformations
to features or latent variables, $\boldsymbol{\chi}_{0}(\mathbf{x})=\left(\chi_{01}(\mathbf{x}),\,...,\,\chi_{0m}(\mathbf{x})\right)^{\top}$
and $\boldsymbol{\chi}_{1}(\mathbf{x})=\left(\chi_{11}(\mathbf{x}),\,...,\,\chi_{1m}(\mathbf{x})\right)^{\top}$,
such that the dynamics in these variables are approximately governed
by the matrix $\mathbf{K}$:
\begin{equation}
\mathbb{E}\left[\boldsymbol{\chi}_{1}\left(\mathbf{x}_{t+\tau}\right)\right]\approx\mathbf{K}^{\top}\mathbb{E}\left[\boldsymbol{\chi}_{0}\left(\mathbf{x}_{t}\right)\right].\label{eq:Kprop}
\end{equation}
This approximation becomes exact in the limit of \textcolor{black}{an
infinitely large set of features ($m\longrightarrow\infty$)} $\boldsymbol{\chi}_{0}$
and $\boldsymbol{\chi}_{1}$, but for a sufficiently large lag time
$\tau$ the approximation can be excellent with low-dimensional feature
transformations, as we will demonstrate below. The expectation values
$\mathbb{E}$ account for stochasticity in the dynamics, such as in
MD, and thus they can be omitted for deterministic dynamical systems
\cite{Mezic_NonlinDyn05_Koopman,TuEtAl_JCD14_ExactDMD,WilliamsKevrekidisRowley_JNS15_EDMD}. 

To illustrate the meaning of Eq. (\ref{eq:Kprop}), consider the example
of $\{\mathbf{x}_{t}\}$ being a discrete-state Markov chain. If we
choose the feature transformation to be indicator functions ($\chi_{0i}=1$
when $\mathbf{x}_{t}=i$ and 0 otherwise, and correspondingly with
$\chi_{1i}$ and $\mathbf{x}_{t+\tau}$), their expectation values
are equal to the probabilities of the chain to be in any given state,
$\mathbf{p}_{t}$ and $\mathbf{p}_{t+\tau}$, and $\mathbf{K}=\mathbf{P}(\tau)$
is equal to the matrix of transition probabilities, i.e. $\mathbf{p}_{t+\tau}=\mathbf{P}^{\top}(\tau)\mathbf{p}_{t}$.
Previous papers on MD kinetics have usually employed a propagator
or transfer operator formulation instead of (\ref{eq:Kprop}) \cite{SchuetteFischerHuisingaDeuflhard_JCompPhys151_146,PrinzEtAl_JCP10_MSM1}.
However, the above formulation is more powerful as it also applies
to nonreversible and non-stationary dynamics, as found for MD of molecules
subject to external force, such as voltage, flow, or radiation \cite{KnochSpeck_NJP15_NoneqMSM,WangSchuette_JCTC15_NoneqMSM}.

A central result of the VAMP theory is that the best finite-dimensional
linear model, i.e. the best approximation in Eq. (\ref{eq:Kprop}),
is found when the subspaces spanned by $\boldsymbol{\chi}_{0}$ and
$\boldsymbol{\chi}_{1}$ are identical to those spanned by the top
$m$ left and right singular functions, respectively, of the so-called
Koopman operator \cite{WuNoe_VAMP}. For an introduction to the Koopman
operator, please refer to \cite{Koopman_PNAS31_Koopman,Mezic_NonlinDyn05_Koopman,KlusEtAl_JNS17_DataDriven}.

How do we choose $\boldsymbol{\chi}_{0}$, $\boldsymbol{\chi}_{1}$
and $\mathbf{K}$ from data? First, suppose we are given some feature
transformation $\boldsymbol{\chi}_{0}$, $\boldsymbol{\chi}_{1}$
and define the following covariance matrices:
\begin{eqnarray}
\mathbf{C}_{00} & = & \mathbb{E}_{t}\left[\boldsymbol{\chi}_{0}\left(\mathbf{x}_{t}\right)\boldsymbol{\chi}_{0}\left(\mathbf{x}_{t}\right)^{\top}\right]\label{eq:C00}\\
\mathbf{C}_{01} & = & \mathbb{E}_{t}\left[\boldsymbol{\chi}_{0}\left(\mathbf{x}_{t}\right)\boldsymbol{\chi}_{1}\left(\mathbf{x}_{t+\tau}\right)^{\top}\right]\label{eq:C01}\\
\mathbf{C}_{11} & = & \mathbb{E}_{t+\tau}\left[\boldsymbol{\chi}_{1}\left(\mathbf{x}_{t+\tau}\right)\boldsymbol{\chi}_{1}\left(\mathbf{x}_{t+\tau}\right)^{\top}\right]\label{eq:C11}
\end{eqnarray}
where $\mathbb{E}_{t}\left[\cdot\right]$ and $\mathbb{E}_{t+\tau}\left[\cdot\right]$
denote the averages that extend over time points and lagged time points
within trajectories, respectively, and across trajectories. Then the
optimal $\mathbf{K}$ that minimizes the least square error $\mathbb{E}_{t}\left[\left\Vert \boldsymbol{\chi}_{1}\left(\mathbf{x}_{t+\tau}\right)-\mathbf{K}^{\top}\boldsymbol{\chi}_{0}\left(\mathbf{x}_{t}\right)\right\Vert ^{2}\right]$
is \cite{WilliamsKevrekidisRowley_JNS15_EDMD,WuNoe_VAMP,HorenkoHartmannSchuetteNoe_PRE07_Langevin}:
\begin{eqnarray}
\mathbf{K} & = & \mathbf{C}_{00}^{-1}\mathbf{C}_{01}.\label{eq:nonreversible_K}
\end{eqnarray}

Now the remaining problem is how to find suitable transformations
$\boldsymbol{\chi}_{0}$, $\boldsymbol{\chi}_{1}$. This problem cannot
be solved by minimizing the least square error above, as is illustrated
by the following example: Suppose we define $\boldsymbol{\chi}_{0}(\mathbf{x})=\boldsymbol{\chi}_{1}(\mathbf{x})=\left(1(\mathbf{x})\right)$,
i.e. we just map the state space to the constant 1 \textendash{} in
this case the least square error is 0 for $\mathbf{K}=\left[1\right]$,
but the model is completely uninformative as all dynamical information
is lost. 

Instead, in order to seek $\boldsymbol{\chi}_{0}$ and $\boldsymbol{\chi}_{1}$
based on available simulation data, we employ the VAMP theorem introduced
in Ref. \cite{WuNoe_VAMP}, that can be equivalently formulated as
the following subspace version:

\textbf{\emph{VAMP variational principle}}\emph{: For any two sets
of linearly independent functions }$\boldsymbol{\chi}_{0}(\mathbf{x})$
and $\boldsymbol{\chi}_{1}(\mathbf{x})$\emph{, let us call
\[
\hat{R}_{2}\left[\boldsymbol{\chi}_{0},\boldsymbol{\chi}_{1}\right]=\left\Vert \mathbf{C}_{00}^{-\frac{1}{2}}\mathbf{C}_{01}\mathbf{C}_{11}^{-\frac{1}{2}}\right\Vert _{F}^{2}
\]
their VAMP-2 score, where $\mathbf{C}_{00},\mathbf{C}_{01},\mathbf{C}_{11}$
are defined by Eqs. (\ref{eq:C00}-\ref{eq:C11}) and $\left\Vert \mathbf{A}\right\Vert _{F}^{2}=n^{-1}\sum_{i,j}A_{ij}^{2}$
is the Frobenius norm of $n\times n$ matrix $\mathbf{A}$. The maximum
value of VAMP-2 score is achieved when the top $m$ left and right
Koopman singular functions belong to $\mathrm{span}(\chi_{01},\,...,\,\chi_{0m})$
and $\mathrm{span}(\chi_{11},\,...,\,\chi_{1m})$ respectively.}

This variational theorem shows that the VAMP-2 score measures the
consistency between subspaces of basis functions and those of dominant
singular functions, and we can therefore optimize $\boldsymbol{\chi}_{0}$
and $\boldsymbol{\chi}_{1}$ via maximizing the VAMP-2 score. In the
special case where the dynamics are reversible with respect to equilibrium
distribution then the theorem above specializes to variational principle
for reversible Markov processes \cite{NoeNueske_MMS13_VariationalApproach,NueskeEtAl_JCTC14_Variational}.
\textcolor{blue}{}

\subsection{Learning the feature transformation using VAMPnets}

\label{subsec:VAMPnet}
\begin{figure}
\centering \includegraphics[width=7cm]{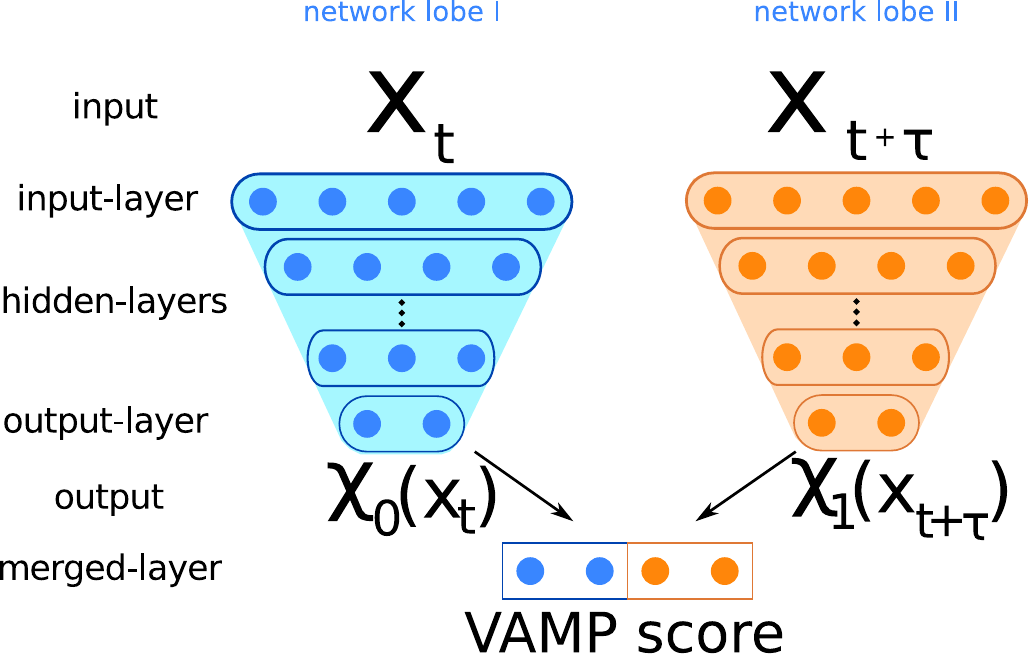} \caption{Scheme of the neural network architecture used. For each time step
$t$ of the simulation trajectory, the coordinates $\mathbf{x}_{t}$
and $\mathbf{x}_{t+\tau}$ are inputs to two deep networks that conduct
a nonlinear dimension reduction. In the present implementation, the
output layer consists of a Softmax classifier. The outputs are then
merged to compute the variational score that is maximized to optimize
the networks. In all present applications the two network lobes are
identical clones, but they can also be trained independently. }
\label{fig:network} 
\end{figure}

Here we employ neural networks to find an optimal
set of basis functions, $\boldsymbol{\chi}_{0}(\mathbf{x})$ and $\boldsymbol{\chi}_{1}(\mathbf{x})$.
Neural networks with at least one hidden layer are universal function
approximators \cite{cybenko1989approximation}, and deep networks
can express strongly nonlinear functions with a fairly few neurons
per layer \cite{eigen2013understanding}. Our networks use VAMP as
a guiding principle and are hence called VAMPnets. VAMPnets consist
of two parallel lobes, each receiving the coordinates of time-lagged
MD configurations $\mathbf{x}_{t}$ and $\mathbf{x}_{t+\tau}$ as
input (Fig. \ref{fig:network}). The lobes have $m$ output nodes
and are trained to learn the transformations $\boldsymbol{\chi}_{0}(\mathbf{x}_{t})$
and $\boldsymbol{\chi}_{1}(\mathbf{x}_{t+\tau})$, respectively. For
a given set of transformations, $\boldsymbol{\chi}_{0}$ and $\boldsymbol{\chi}_{1}$,
we pass a batch of training data through the network and compute the
training VAMP-2 score of our choice. VAMPnets bear similarities with
auto-encoders \cite{RanzatoLeCun_NIPS06_Autoencoder,BengioEtAl_NIPS07_Autoencoder}
using a time-delay embedding and are closely related to deep Canonical
Covariance Analysis (CCA) \cite{DCCA}. VAMPnets are identical to
deep CCA with time-delay embedding when using the VAMP-1 score discussed
in \cite{WuNoe_VAMP}, however the VAMP-2 score has easier-to-handle
gradients and is more suitable for time series data, due to its direct
relation to the Koopman approximation error \cite{WuNoe_VAMP}. 

The first left and right singular functions of the Koopman operator
are always equal to the constant function $\mathbbm1(\mathbf{x})\equiv1$
\cite{WuNoe_VAMP}. We can thus add $\mathbbm1$ to basis functions
and train the network by maximizing
\begin{equation}
\hat{R}_{2}\left[\left(\begin{array}{c}
\mathbbm1\\
\boldsymbol{\chi}_{0}
\end{array}\right),\left(\begin{array}{c}
\mathbbm1\\
\boldsymbol{\chi}_{1}
\end{array}\right)\right]=\left\Vert \bar{\mathbf{C}}_{00}^{-\frac{1}{2}}\bar{\mathbf{C}}_{01}\bar{\mathbf{C}}_{11}^{-\frac{1}{2}}\right\Vert _{F}^{2}+1,\label{eq:VAMP2-score}
\end{equation}
where $\bar{\mathbf{C}}_{00},\bar{\mathbf{C}}_{01},\bar{\mathbf{C}}_{11}$
are mean-free covariances of the feature-transformed coordinates:
\begin{eqnarray}
\bar{\mathbf{C}}_{00} & = & (T-1)^{-1}\bar{\mathbf{X}}\bar{\mathbf{X}}^{\top}\label{eq:C00-bar}\\
\bar{\mathbf{C}}_{01} & = & (T-1)^{-1}\bar{\mathbf{X}}\bar{\mathbf{Y}}^{\top}\\
\bar{\mathbf{C}}_{11} & = & (T-1)^{-1}\bar{\mathbf{Y}}\bar{\mathbf{Y}}^{\top}.\label{eq:C11-bar}
\end{eqnarray}
Here we have defined the matrices $\mathbf{X}=[X_{ij}]=\chi_{0i}(\mathbf{x}_{j})\in\mathbb{R}^{m\times T}$
and $\mathbf{Y}=[Y_{ij}]=\chi_{1i}(\mathbf{x}_{j+\tau})\in\mathbb{R}^{m\times T}$
with $\{(\mathbf{x}_{j},\mathbf{x}_{j+\tau})\}_{j=1}^{T}$ representing
all available transition pairs, and their mean-free versions $\bar{\mathbf{X}}=\mathbf{X}-T^{-1}\mathbf{X}\mathbf{1}$,
$\bar{\mathbf{Y}}=\mathbf{Y}-T^{-1}\mathbf{Y}\mathbf{1}$. The gradients
of $\hat{R}_{2}$ are given by: 
\begin{align}
\nabla_{\mathbf{X}}\hat{R}_{2} & =\frac{2}{T-1}\bar{\mathbf{C}}_{00}^{-1}\bar{\mathbf{C}}_{01}\bar{\mathbf{C}}_{11}^{-1}\left(\bar{\mathbf{Y}}-\bar{\mathbf{C}}_{01}^{\top}\bar{\mathbf{C}}_{00}^{-1}\bar{\mathbf{X}}\right)\label{eq:gradX}\\
\nabla_{\mathbf{Y}}\hat{R}_{2} & =\frac{2}{T-1}\bar{\mathbf{C}}_{11}^{-1}\bar{\mathbf{C}}_{01}^{\top}\bar{\mathbf{C}}_{00}^{-1}\left(\bar{\mathbf{X}}-\bar{\mathbf{C}}_{01}\bar{\mathbf{C}}_{11}^{-1}\bar{\mathbf{Y}}\right)\label{eq:gradY}
\end{align}
and are back-propagated to train the two network lobes. See Supplementary
Note \textcolor{black}{1} for derivations of Eqs. (\ref{eq:VAMP2-score},\ref{eq:gradX},\ref{eq:gradY}).

For simplicity of interpretation we may just use a unique basis set
$\boldsymbol{\chi}=\boldsymbol{\chi}_{0}=\boldsymbol{\chi}_{1}$.
Even when using two different basis sets would be meaningful, we can
unify them by simply defining $\boldsymbol{\chi}=(\boldsymbol{\chi}_{0},\boldsymbol{\chi}_{1})^{\top}$.
In this case, we clone the lobes of the network and train them using
the total gradient $\nabla\hat{R}_{2}=\nabla_{\mathbf{X}}\hat{R}_{2}+\nabla_{\mathbf{Y}}\hat{R}_{2}$. 

After training, we asses the quality of the learned features and select
hyper-parameters (e.g. network size) while avoiding overfitting using
the VAMP-2 validation score 
\begin{equation}
\hat{R}_{2}^{\mathrm{val}}=\left\Vert \left(\bar{\mathbf{C}}_{00}^{\mathrm{val}}\right)^{-\frac{1}{2}}\bar{\mathbf{C}}_{01}^{\mathrm{val}}\left(\bar{\mathbf{C}}_{11}^{\mathrm{val}}\right)^{-\frac{1}{2}}\right\Vert _{F}^{2}+1,
\end{equation}
where $\bar{\mathbf{C}}_{00}^{\mathrm{val}},\bar{\mathbf{C}}_{01}^{\mathrm{val}},\bar{\mathbf{C}}_{11}^{\mathrm{val}}$
are mean-free covariance matrices computed from a validation data
set not used during the training.

\subsection{Dynamical model and validation}

\textcolor{black}{The direct estimate of the time-lagged covariance
matrix $\mathbf{C}_{01}$ is generally nonsymmetric. Hence the Koopman
model or Markov state model $\mathbf{K}$ given by Eq. (\ref{eq:nonreversible_K})
is typically not time-reversible \cite{WuEtAl_JCP17_VariationalKoopman}.
In MD, it is often desirable to obtain a time-reversible kinetic model
\textendash{} see \cite{TrendelkampSchroerEtAl_InPrep_revMSM} for
a detailed discussion. To enforce reversibility, $\mathbf{K}$ can
be reweighted as described in \cite{WuEtAl_JCP17_VariationalKoopman}
and implemented in PyEMMA \cite{SchererEtAl_JCTC15_EMMA2}. The present
results do not depend on enforcing reversibility, as classical analyses
such as PCCA+ \cite{RoeblitzWeber_AdvDataAnalClassif13_PCCA++} are
avoided based on the VAMPnet structure itself.}

Since $\mathbf{K}$ is a Markovian model, it is expected to fulfill
the Chapman-Kolmogorov (CK) equation:
\begin{equation}
\mathbf{K}(n\tau)=\mathbf{K}^{n}(\tau),\label{eq:CK}
\end{equation}
for any value of $n\ge1$, where $\mathbf{K}(\tau)$ and $\mathbf{K}(n\tau)$
indicate the models estimated at a lag time of $\tau$ and $n\tau$,
respectively. However, since any Markovian model of MD can be only
approximate \cite{SarichNoeSchuette_MMS09_MSMerror,PrinzEtAl_JCP10_MSM1},
Eq. (\ref{eq:CK}) can only be fulfilled approximately, and the relevant
test is whether it holds within statistical uncertainty. We construct
two tests based on Eq. (\ref{eq:CK}): In order to select a suitable
dynamical model, we will proceed as for Markov state models by conducting
an eigenvalue decomposition for every estimated Koopman matrix, $\mathbf{K}(\tau)\mathbf{r}_{i}=\mathbf{r}_{i}\lambda_{i}(\tau)$,
and computing the implied timescales \cite{SwopePiteraSuits_JPCB108_6571}
as a function of lag time:
\begin{equation}
t_{i}(\tau)=-\frac{\tau}{\ln|\lambda_{i}(\tau)|},\label{eq:its}
\end{equation}
We chose a value $\tau$, where $t_{i}(\tau)$ are approximately constant
in $\tau$. After having chosen $\tau$, we will test whether Eq.
(\ref{eq:CK}) holds within statistical uncertainty \cite{NoeSchuetteReichWeikl_PNAS09_TPT}.
For both the implied timescales and the CK-test we proceed as follows:
train the neural network at a fixed lag time $\tau^{*}$, thus obtaining
the network transformation $\boldsymbol{\chi}$, and then compute
Eq. (\ref{eq:CK}) or Eq. (\ref{eq:its}) for different values of
$\tau$ with a fixed transformation $\boldsymbol{\chi}$. Finally,
the approximation of the $i$th eigenfunction is given by
\begin{equation}
\hat{\psi}_{i}^{e}(\mathbf{x})=\sum_{j}r_{ij}\chi_{j}(\mathbf{x}).\label{eq:K_eigenfunction}
\end{equation}
If dynamics are reversible, the singular value decomposition and eigenvalue
decomposition are identical, i.e. $\sigma_{i}=\lambda_{i}$ and $\psi_{i}=\psi_{i}^{e}$.

\subsection{Network architecture and training}

We use VAMPnets to learn molecular kinetics from simulation data of
a range of model systems. While any neural network architecture can
be employed inside the VAMPnet lobes, we chose the following setup
for our applications: \textcolor{black}{the two network lobes are
identical clones, i.e. $\boldsymbol{\chi}_{0}\equiv\boldsymbol{\chi}_{1}$,
and consist}\textcolor{blue}{{} }of fully connected networks. In most
cases, the networks have less output than input nodes, i.e. the network
conducts a dimension reduction. In order to divide the work equally
between network layers, we reduce the number of nodes from each layer
to the next by a constant factor. Thus, the network architecture is
defined by two parameters: the depth $d$ and the number of output
nodes $n_{\mathrm{out}}$. All hidden layers employ Rectified Linear
Units (ReLU) \cite{hahnloser1998rectified,nair2010rectified}.

Here, we build the output layer with Softmax output nodes, i.e. $\chi_{i}(\mathbf{x})\ge0$
for all $i$ and $\sum_{i}\chi_{i}(\mathbf{x})=1$. Therefore, the
activation of an output node can be interpreted as a probability to
be in state $i$. As a result, the network effectively performs featurization,
dimension reduction and finally a fuzzy clustering to metastable states,
and the $\mathbf{K}(\tau)$ matrix computed from the network-transformed
data is the transition matrix of a fuzzy MSM \cite{WuNoe_JCP15_GMTM,WeberFackeldeySchuette_JCP17_SetfreeMSM,HarriganPande_bioRxiv17_LandmarkTICA}.
Consequently, Eq. (\ref{eq:Kprop}) propagates probability distributions
in time. 

The networks were trained with pairs of MD configurations $(\mathbf{x}_{t},\,\mathbf{x}_{t+\tau})$
using the Adam stochastic gradient descent method \cite{adam}. \textcolor{black}{For
each result we repeated 100 training runs, each of which with a randomly
chosen 90\%/10\% division of the data to training and validation data.}\textcolor{blue}{{}
}See Methods section for details on network architecture, training
and choice of hyper-parameters.

\subsection{Asymmetric double well potential}

\begin{figure}
\centering{} \includegraphics[width=0.8\columnwidth]{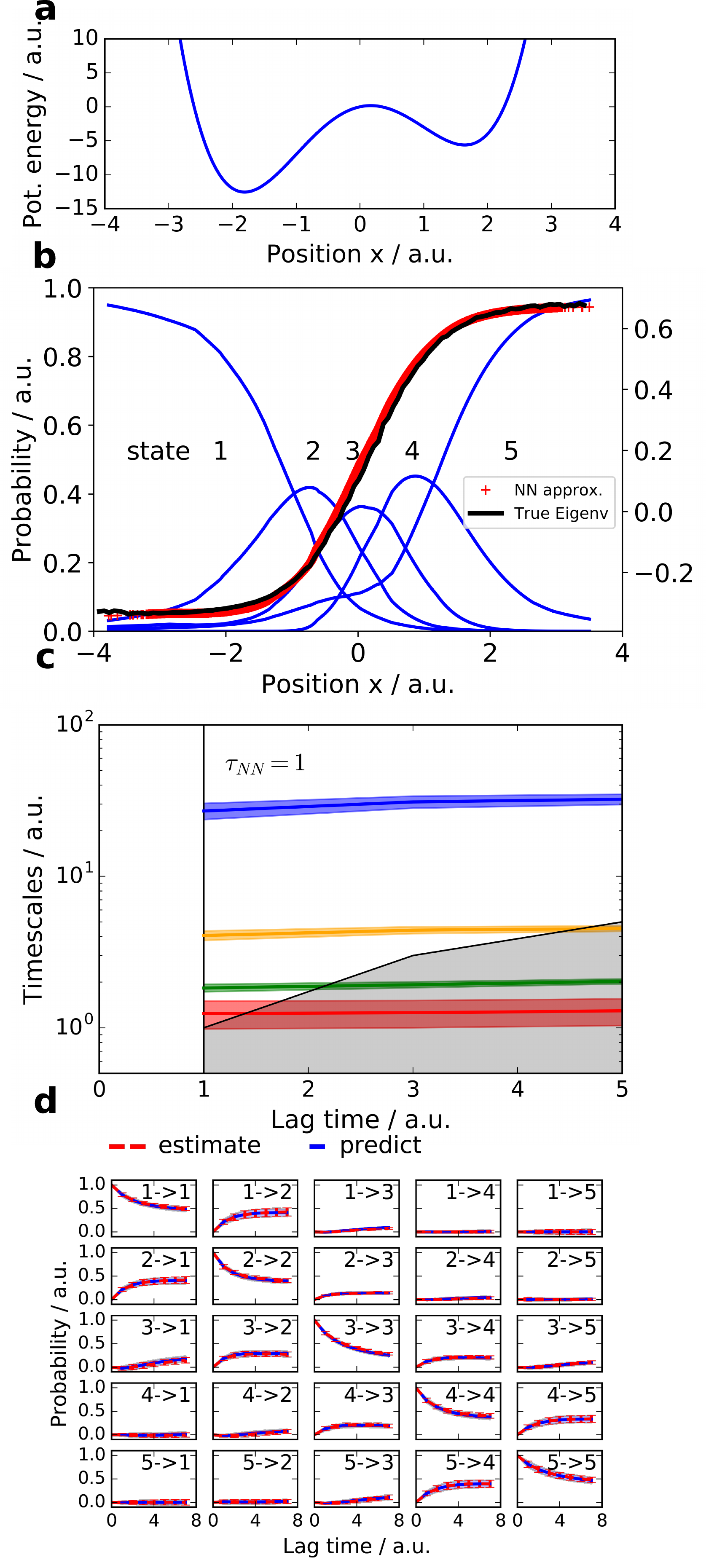} \caption{Approximation of the slow transition in a bistable potential. (\textbf{a})
Potential energy function $U(x)=x^{4}-6x^{2}+2x$. (\textbf{b}) Eigenvector
of the slowest process calculated \textcolor{black}{by direct numerical
approximation (black) }and approximated by a VAMPnet with five output
nodes (red). Activation of the five Softmax output nodes define the
state membership probabilities (blue). (\textbf{c}) Relaxation timescales
computed from the Koopman model using the VAMPnet transformation.
(\textbf{d}) Chapman-Kolmogorov test comparing long-time predictions
of the Koopman model estimated at $\tau=1$ and estimates at longer
lag times. Panels (c) and (d) report 95\% confidence interval error
bars over 100 training runs.}
\label{fig:1dpot} 
\end{figure}
We first model the kinetics of a bistable one-dimensional process,
simulated by Brownian dynamics (see Methods) in an asymmetric double-well
potential (Fig. \ref{fig:1dpot}a). A trajectory of $50,000$ time
steps is generated. Three-layer VAMPnets are set up with 1-5-10-5
nodes in each lobe. The single input node of each lobe is given the
current and time-lagged mean-free $x$ coordinate of the system, i.e.
$x_{t}-\mu_{1}$ and $x_{t+\tau}-\mu_{2}$, where $\mu_{1}$ and $\mu_{2}$
are the respective means, and $\tau=1$ is used. The network maps
to five Softmax output nodes that we will refer to as \emph{states},
as the network performs a fuzzy discretization by mapping the input
configurations to the output activations. The network is trained by
using the VAMP-2 score with the four largest singular values.

The network learns to place the output states in a way to resolve
the transition region best (Fig. \ref{fig:1dpot}b), which is known
to be important for the accuracy of a Markov state model \cite{SarichNoeSchuette_MMS09_MSMerror,PrinzEtAl_JCP10_MSM1}.
This placement minimizes the Koopman approximation error, as seen
by comparing the dominant Koopman eigenfunction (Eq. \ref{eq:K_eigenfunction})
with \textcolor{black}{a direct numerical approximation of the true
eigenfunction obtained by a transition matrix computed for a direct
uniform 200-state discretization of the $x$ axis \textendash{} see
\cite{PrinzEtAl_JCP10_MSM1} for details}. The implied timescale and
Chapman-Kolmogorov tests (Eqs. \ref{eq:its} and \ref{eq:CK}) confirm
that the kinetic model learned by the VAMPnet successfully predicts
the long-time kinetics (Fig. \ref{fig:1dpot} c,d).

\subsection{Protein folding model}

While the first example was one-dimensional we now test if VAMPnets
are able to learn reaction coordinates that are nonlinear functions
of a multi-dimensional configuration space. For this, we simulate
a $100,000$ time step Brownian dynamics trajectory (Eq. \ref{eq:Brown})
using the simple protein folding model defined by the potential energy
function (Supplementary Fig. 1 a):
\[
U(r)=\begin{cases}
-2.5\,(r-3)^{2} & r<3\\
0.5\,(r-3)^{3}-(r-3)^{2} & r\ge3
\end{cases}
\]
The system has a five-dimensional configuration space, $\mathbf{x}\in\mathbb{R}^{5}$,
however the energy only depends on the norm of the vector $r=|\mathbf{x}|$.
While small values of $r$ are energetically favorable, large values
of $r$ are entropically favorable as the number of configurations
available on a five-dimensional hypersphere grows dramatically with
$r$. Thus, the dynamics are bistable along the reaction coordinate
$r$. Four-layer network lobes with 5-32-16-8-2 nodes each were employed
and trained to maximize the VAMP-2 score involving the largest nontrivial
singular value.

The two output nodes successfully identify the folded and the unfolded
states, and use intermediate memberships for the intersecting transition
region (Supplementary Fig. 1 b). The network excellently approximates
the Koopman eigenfunction of the folding process, as apparent from
the comparison of the values of the network eigenfunction computed
by Eq. (\ref{eq:K_eigenfunction}) with the eigenvector computed from
a high-resolution MSM built on the $r$ coordinate (Supplementary
Fig. 1 b). This demonstrates that the network can learn the nonlinear
reaction coordinate mapping $r=|\mathbf{x}|$ based only on maximizing
the variational score \ref{eq:VAMP2-score}. Furthermore, the implied
timescales and the CK-test indicate that the network model predicts
the long-time kinetics almost perfectly (Supplementary Fig. 1 c,d).

\subsection{Alanine dipeptide}

\label{subsec:ala2}As a next level, VAMPnets are used to learn the
kinetics of alanine dipeptide from simulation data. It is known that
the $\phi$ and $\psi$ backbone torsion angles are the most important
reaction coordinates that separate the metastable states of alanine
dipeptide, however, our networks only receive Cartesian coordinates
as an input, and are thus forced to learn both the nonlinear transformation
to the torsion angle space and an optimal cluster discretization within
this space, in order to obtain an accurate kinetic model.

\begin{figure}
\centering{}%
\begin{minipage}[t][1\totalheight][b]{0.5\columnwidth}%
\begin{center}
\includegraphics[width=1\textwidth]{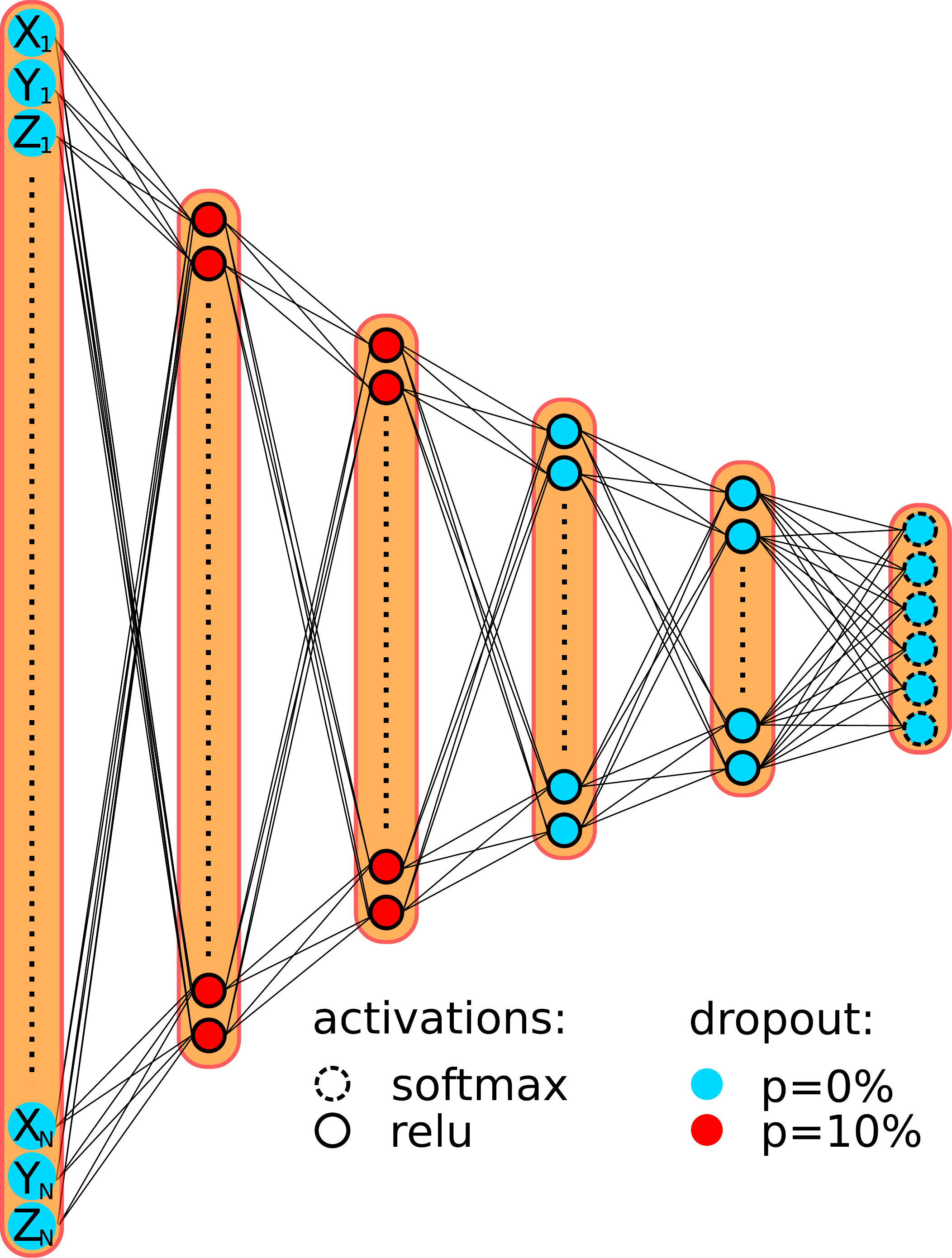}
\par\end{center}%
\end{minipage}\hfill%
\begin{minipage}[t][1\totalheight][b]{0.45\columnwidth}%
\begin{center}
\caption{\label{fig:ala_network}Representative structure of one lobe of the
VAMPnet used for alanine dipeptide. Here, the five-layer network with
six output states used for the results shown in Fig. \ref{fig:ala1}
is shown. Layers are fully connected, have 30-22-16-12-9-6 nodes,
and use dropout in the first two hidden layers. All hidden neurons
use ReLu activation functions, while the output layer uses Softmax
activation function in order to achieve a fuzzy discretization of
state space.}
\par\end{center}%
\end{minipage} 
\end{figure}

A $250$ nanosecond MD trajectory generated in Ref. \cite{NueskeEtAl_JCP17_OOMMSM}
(MD setup described there) serves as a dataset. The solute coordinates
were stored every picosecond, resulting in $250,000$ configurations
that are all aligned on the first frame using minimal RMSD fit to
remove global translation and rotation. Each network lobe uses the
three-dimensional coordinates of the 10 heavy atoms as input, $(x_{1},y_{1},z_{1},\,...,\,x_{10},y_{10},z_{10})$,
and the network is trained using time lag $\tau=\unit{40}\,\mathrm{ps}$.
Different numbers of output states and layer depths are considered,
employing the layer sizing scheme described in the Methods section
(see Fig. \ref{fig:ala_network} for an example). 

\begin{figure}
\centering{}\includegraphics[width=0.7\columnwidth]{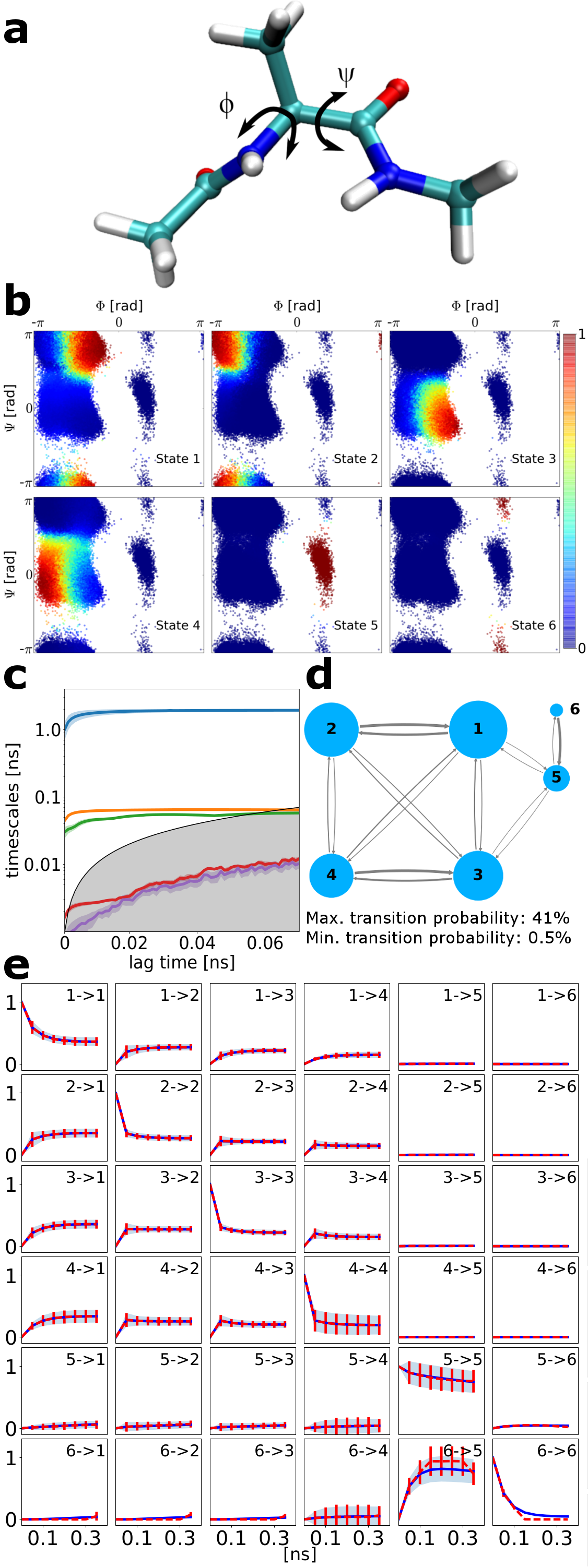} \caption{VAMPnet kinetic model of alanine dipeptide. (\textbf{a}) Structure
of alanine dipeptide. The main coordinates describing the slow transitions
are the backbone torsion angles $\phi$ and $\psi$, however the neural
network inputs are only the Cartesian coordinates of heavy atoms.
(\textbf{b}) Assignment of all simulated molecular coordinates, plotted
as a function of $\phi$ and $\psi$, to the six Softmax output states.
Color corresponds to activation of the respective output neuron, indicating
the membership probability to the associated metastable state. (\textbf{c})
Relaxation timescales computed from the Koopman model using the neural
network transformation. (\textbf{d}) Representation of the transition
probabilities matrix of the Koopman model; transitions with a probability
lower than 0.5\% have been omitted. (\textbf{e}) Chapman-Kolmogorov
test comparing long-time predictions of the Koopman model estimated
at $\tau=50\,ps$ and estimates at longer lag times. Panels (c) and
(e) report 95\% confidence interval error bars over 100 training runs
excluding failed runs (see text).}
\label{fig:ala1} 
\end{figure}

A VAMPnet with six output states learns a discretization in six metastable
sets corresponding to the free energy minima of the $\phi/\psi$ space
(Fig. \ref{fig:ala1}b). The implied timescales indicate that given
the coordinate transformation found by the network, the two slowest
timescales are converged at lag time $\tau=50\,\mathrm{ps}$ or larger
(Fig. \ref{fig:ala1}c). Thus we estimated a Koopman model at $\tau=50\,\mathrm{ps}$,
whose Markov transition probability matrix is depicted in Fig. \ref{fig:ala1}d.
Note that transition probabilities between state pairs $1\leftrightarrow4$
and $2\leftrightarrow3$ are important for the correct kinetics at
$\tau=50\,\mathrm{ps}$, but the actual trajectories typically pass
via the directly adjacent intermediate states. The model performs
excellently in the CK-Test (Fig. \ref{fig:ala1}e). 

\subsection{Choice of lag time, network depth and number of output states}

We studied the success probability of optimizing a VAMPnet with\textcolor{black}{{}
six output states as a function of the lag time $\tau$ by conducting
200 optimization runs. Success was defined as resolving the three
slowest processes by finding three slowest timescale higher than 0.2,
0.4 and 1 ns, respectively. Note that the results shown in Fig. \ref{fig:ala1}
are reported for successful runs in this definition. There is a range
of $\tau$ values from 4 to 32 picoseconds where the training succeeds
with a significant probability (}Supplementary Fig. \textcolor{black}{2
a). However, even in this range the success rate is still below 40
\%, which is mainly due to the fact that many runs fail to find the
rarely occurring third-slowest process that corresponds to the $\psi$
transition of the positive $\phi$ range (Fig. \ref{fig:ala1}b, state
5 and 6). }

The breakdown of optimization success for small and large lag times
can be most easily explained by the eigenvalue decomposition of Markov
propagators \cite{PrinzEtAl_JCP10_MSM1}. When the lag time exceeds
the timescale of a process, the amplitude of this process becomes
negligible, making it hard to fit given noisy data. At short lag times,
many processes have large \textcolor{black}{eigenvalues, which increases
the search space of the neural network and appears to increase the
probability of getting stuck in suboptimal maxima of the training
score.}

\textcolor{black}{We have also studied the success probability, as
defined above, as a function of network depth. Deeper networks can
represent more complex functions. Also, since the networks defined
here reduce the input dimension to the output dimension by a constant
factor per layer, deeper networks perform a less radical dimension
reduction per layer. On the other hand, deeper networks are more difficult
to train. As seen in}Supplementary Fig. \textcolor{black}{2 b, a high
success rate is found for four to seven layers. }

\begin{figure*}[t]
\centering{}\includegraphics[width=0.9\textwidth]{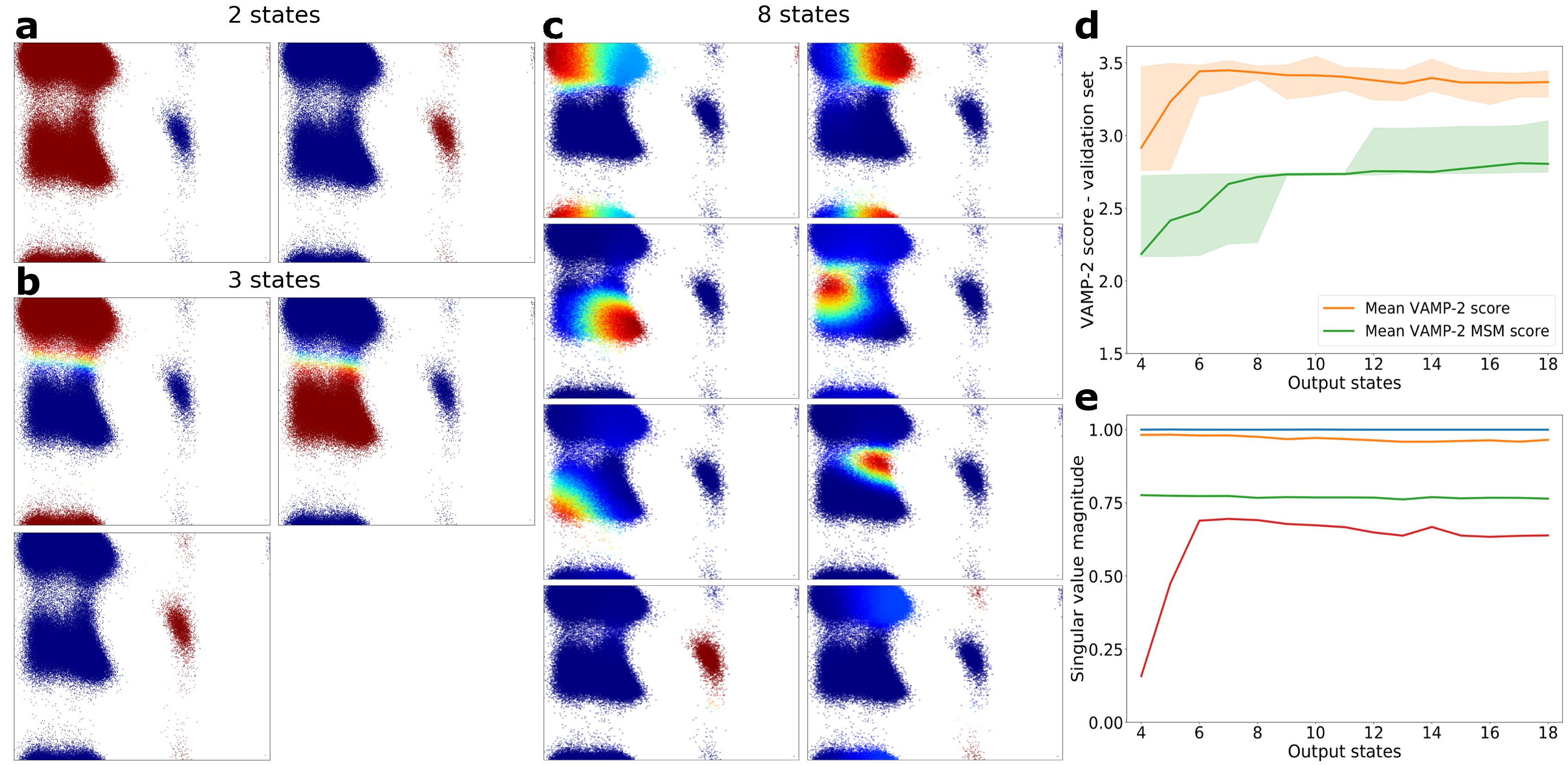} \caption{\label{fig:Ala_hierarchical_decomp}Kinetic model of alanine dipeptide
as a function of the number of output states. (\textbf{a-c}) Assignment
of input coordinates, plotted as a function of $\phi$ and $\psi$,
to two, three, and eight output states. C\textcolor{black}{olor corresponds
to activation of the respective output neuron, indicating the membership
probability to this state (see \ref{fig:ala1} b). (}\textbf{\textcolor{black}{d}}\textcolor{black}{)
Comparison of VAMPnet and MSM performance as a function of the number
of output states / MSM states. Mean VAMP-2 score and 95\% confidence
interval from 100 runs are shown. (}\textbf{\textcolor{black}{e}}\textcolor{black}{)
Mean squared values of the four largest singular values that make
up the VAMPnets score plotted in panel} (d).}
 
\end{figure*}

Next, we studied the dependency of the network-based discretization
as a function of the number of output nodes (Fig. \ref{fig:Ala_hierarchical_decomp}a-c).
With two output states, the network separates the state space at the
slowest transition between negative and positive values of the $\phi$
angle (Fig. \ref{fig:Ala_hierarchical_decomp}a). The result with
three output nodes keeps the same separation and additionally distinguishes
between the $\alpha$ and $\beta$ regions of the Ramachandran plot,
i.e. small and large values of the $\psi$ angle (Fig. \ref{fig:Ala_hierarchical_decomp}b).
For higher number of output states, finer discretizations and smaller
interconversion timescales are found, until the network starts discretizing
the transition regions, such as the two transition states between
the $\alpha$ and $\beta$ regions along the $\psi$ angle (Fig. \ref{fig:Ala_hierarchical_decomp}c).
We chose the lag time depending on the number of output nodes of the
network, using $\tau=200\,\mathrm{ps}$ for two output nodes, $\tau=60\,\mathrm{ps}$
for three output nodes, and $\tau=1\,\mathrm{ps}$ for eight output
nodes. 

An network output with $k$ Softmax neurons describes a $(k-1)$-dimensional
feature space as the Softmax normalization removes one degree of freedom.
Thus, to resolve $k-1$ relaxation timescales, at least $k$ output
nodes or metastable states are required. However, the network quality
can improve when given more degrees of freedom in order to approximate
the dominant singular functions accurately. Indeed, the best scores
using $k=4$ singular values (3 nontrivial singular values) are achieved
when using at least \textcolor{black}{six} output states that separate
each of the six metastable states in the Ramachandran plane (Fig.
\ref{fig:Ala_hierarchical_decomp}d-e). 

\textcolor{black}{For comparison, we investigated how a standard MSM
would perform as a function of the number of states (Fig. \ref{fig:Ala_hierarchical_decomp}d).
For a fair comparison, the MSMs also used Cartesian coordinates as
an input, but then employed a state-of-the-art procedure using a kinetic
map transformation that preserves 95\% of the cumulative kinetic variance
\cite{NoeClementi_JCTC15_KineticMap}, followed by $k$-means clustering,
where the parameter $k$ is varied. It is seen that the MSM VAMP-2
scores obtained by this procedure is significantly worse than by VAMPnets
when less than 20 states are employed. Clearly, MSMs will succeed
when sufficiently many states are used, but in order to obtain an
interpretable model those states must again be coarse-grained onto
a fewer-state model, while VAMPnets directly produce an accurate model
with few-states.}

\subsection{VAMPnets learn to transform Cartesian to torsion coordinates}

The results above indicate that the VAMPnet has implicitly learned
the feature transformation from Cartesian coordinates to backbone
torsions. In order to probe this ability more explicitly, we trained
a network with 30-10-3-3-2-5 layers, i.e. including a bottleneck of
two nodes before the output layer. We find that the activation of
the two bottleneck nodes correlates excellently with the $\phi$ and
$\psi$ torsion angles that were not presented to the network (Pearson
correlation coefficients of $0.95$ and $0.92$, respectively, Supplementary
Fig. \textcolor{black}{3 }a,b). To visualize the internal representation
that the network learns, we color data samples depending on the free
energy minima in the $\phi/\psi$ space they belong to (Supplementary
Fig. \textcolor{black}{3 }c), and then show where these samples end
up in the space of the bottleneck node activations (Supplementary
Fig. \textcolor{black}{3 }d). It is apparent that the network learns
a representation of the Ramachandran plot \textendash{} The four free
energy minima at small $\phi$ values ($\alpha_{R}$ and $\beta$
areas) are represented as contiguous clusters with the correct connectivity,
and are well separated from states with large $\phi$ values ($\alpha_{L}$
area). The network fails to separate the two substates in the large
$\phi$ value range well, which explains the frequent failure to find
the corresponding transition process and the third-largest relaxation
timescale.

\subsection{NTL9 Protein folding dynamics}

In order to proceed to a higher-dimensional problem, we analyze the
kinetics of an all-atom protein folding simulation of the NTL9 protein
generated by the Anton supercomputer \cite{LindorffLarsenEtAl_Science11_AntonFolding}.
A five-layer VAMPnet was trained at lag time $\tau=\unit[10]{ns}$
using $111,000$ time steps, uniformly sampled from a $\unit[1.11]{ms}$
trajectory. Since NTL9 is folding and unfolding, there is no unique
reference structure to align Cartesian coordinates to \textendash{}
hence we use internal coordinates as a network input. We computed
the nearest-neighbor heavy-atom distance, $d_{ij}$ for all non-redundant
pairs of residues $i$ and $j$ and transformed them into contact
maps using the definition $c_{ij}=\exp(-d_{ij})$, resulting in \textbf{$666$}
input nodes. 

\begin{figure*}
\begin{minipage}[t][1\totalheight][b]{0.75\textwidth}%
\includegraphics[width=1\textwidth]{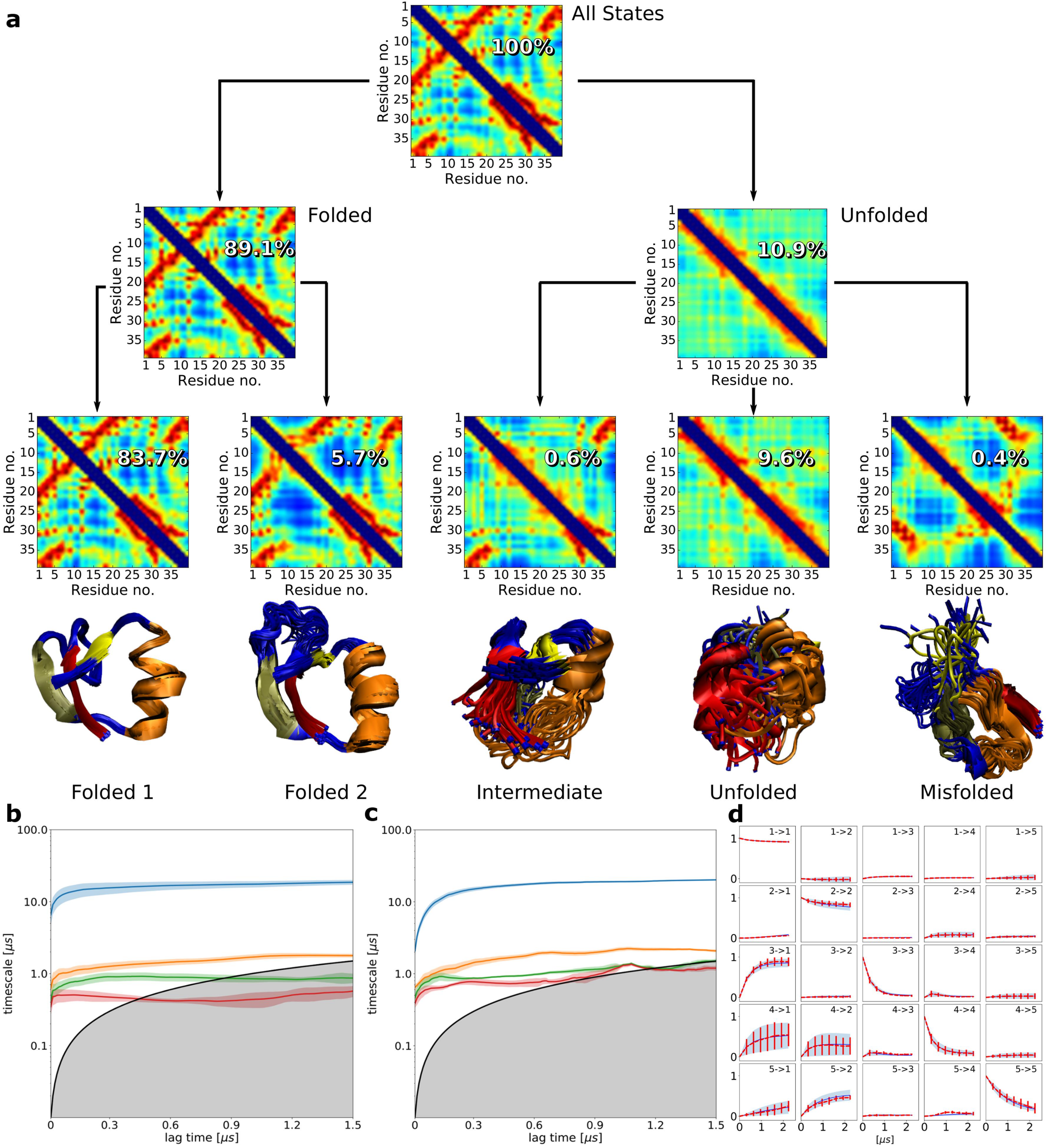}%
\end{minipage}\hfill%
\begin{minipage}[t][1\totalheight][b]{0.22\textwidth}%
\caption{\label{fig:NTL9}VAMPnet results of NTL9 folding kinetics. (a) Hierarchical
decomposition of the NTL9 protein state space by a network with two
and five output nodes. Mean contact maps are shown for all MD samples
grouped by the network, along with the fraction of samples in that
group. 3D structures are shown for the five-state decomposition, residues
involved in $\alpha$-helices or $\beta$-sheets in the folded state
are colored identically across the different states. (b) Relaxation
timescales computed from the Koopman model approximated using the
transformation applied by a neural network with five output nodes.
(c) Relaxation timescales from a Markov state model computed from
a TICA transformation of the contact maps, followed by k-means clustering
with $k=40$. (d) Chapman-Kolmogorov test comparing long-time predictions
of the Koopman model estimated at $\tau=320ns$ and estimates at longer
lag times. Panels (b), (c) and (d) report 95\% confidence interval
error bars over 100 training runs.}
\end{minipage}
\end{figure*}

Again, the network performs a hierarchical decomposition of the molecular
configuration space when increasing the number of output nodes. Fig.
\ref{fig:NTL9}a shows the decomposition of state space for two and
five output nodes, and the corresponding mean contact maps and state
probabilities. With two output nodes, the network finds the folded
and unfolded state that are separated by the slowest transition process
(Fig. \ref{fig:NTL9}a, middle row). With five output states, the
folded state is decomposed into a stable and well-defined fully folded
substate and a less stable, more flexible substate that is missing
some of the tertiary contacts compared to the fully folded substate.
The unfolded substate decomposes into three substates, one of them
largely unstructured, a second one with residual structure, thus forming
a folding intermediate, and a mis-folded state with an entirely different
fold including a non-native $\beta$-sheet. 

The relaxation timescales found by a five-state VAMPnet model are
\emph{en par} with those found by a 40-state MSM using state-of-the-art
estimation methods (Fig. \ref{fig:NTL9}b-c). However, the fact that
only five states are required in the VAMPnet model makes it easier
to interpret and analyze. Additionally, the CK-test indicates excellent
agreement between long-time predictions and direct estimates.

\section*{Discussion}

We have introduced a deep learning framework for molecular kinetics,
called VAMPnet. Data-driven learning of molecular kinetics is usually
done by shallow learning structures, such as TICA and MSMs. However,
the processing pipeline, typically consisting of featurization, dimension
reduction, MSM estimation and MSM coarse-graining is, in principle,
a hand-crafted deep learning structure. Here we propose to replace
the entire pipeline by a deep neural network that learns optimal feature
transformations, dimension reduction and, if desired, maps the MD
time-steps to a fuzzy clustering. The key to optimize the network
is the VAMP variational approach which defines scores by which learning
structures can be optimized to learn models of both equilibrium and
non-equilibrium MD.

Although MSM-based kinetic modeling has been refined over more than
a decade, VAMPnets perform competitively or superior in our examples.
In particular, they perform extremely well in the Chapman-Kolmogorov
test that validates the long-time prediction of the model. VAMPnets
have a number of advantages over models based on MSM pipelines: (i)
They may be overall more optimal, because featurization, dimension
reduction and clustering are not explicitly separate processing steps.
(ii) When using Softmax output nodes, the VAMPnet performs a fuzzy
clustering of the MD structures fed into the network and constructs
a fuzzy MSM, which is readily interpretable in terms of transition
probabilities between metastable states. In contrast to other MSM
coarse-graining techniques it is thus not necessary to accept reduction
in model quality in order to obtain a few-state MSM, but such a coarse-grained
model is seamlessly learned within the same learning structure. (iii)
VAMPnets require less expertise to train than an MSM-based processing
pipelines, and the formulation of the molecular kinetics as a neural
network learning problem enables us to exploit an arsenal of highly
developed and optimized tools in learning softwares such as \emph{tensorflow},
\emph{theano} or \emph{keras}.

Despite their benefits, VAMPnets still miss many of the benefits that
come with extensions developed for MSM approach. This includes multi-ensemble
Markov models that are superior to single-conventional MSMs in terms
of sampling rare events by combining data from multiple ensembles
\cite{WuEtAL_PNAS16_TRAM,WuMeyRostaNoe_JCP14_dTRAM,ChoderaEtAl_JCP11_DynamicalReweighting,PrinzEtAl_JCP11_Reweighting,RostaHummer_DHAM,MeyWuNoe_xTRAM},
Augmented Markov models that combine simulation data with experimental
observation data \cite{OlssonEtAl_PNAS17_AugmentedMarkovModels},
and statistical error estimators developed for MSMs \cite{Singhal_JCP07,Noe_JCP08_TSampling,ChoderaNoe_JCP09_MSMstatisticsII}.
Since these methods explicitly use the MSM likelihood, it is currently
unclear, how they could be implemented in a deep learning structure
such as a VAMPnet. Extending VAMPnets towards these special capabilities
is a challenge for future studies.

Finally, a remaining concern is that the optimization of VAMPnets
can get stuck in suboptimal local maxima. In other applications of
network-based learning, a working knowledge has been established which
type of network implementation and learning algorithm are most suitable
for robust and reproducible learning. For example, it is conceivable
that the VAMPnet lobes may benefit from convolutional filters \cite{LeCun_ProcIEEE89_ConvNets}
or different types of transfer functions.\textcolor{blue}{{} }\textcolor{black}{Suitably
chosen convolutions, as in \cite{SchuettEtAl_MoleculeNet2017} may
also lead to learned feature transformations that are transferable
within a given class of molecules.}

\section*{Methods}

\paragraph*{Neural network structure}

Each network lobe in Fig. \ref{fig:network} has a number of input
nodes given by the data dimension. According to the VAMP variational
principle (Sec. \ref{subsec:VAMP-principle}), the output dimension
must be at least equal to the number of Koopman singular functions
that we want to approximate, i.e. equal to $k$ used in the score
function $\hat{R}_{2}$. In most applications, the number of input
nodes exceeds the number of output nodes, i.e. the network conducts
a dimension reduction. Here, we keep the dimension reduction from
layer $i$ with $n_{i}$ nodes to layer $i+1$ with $n_{i+1}$ nodes
constant:
\begin{equation}
\frac{n_{i}}{n_{i+1}}=\left(\frac{n_{\mathrm{in}}}{n_{\mathrm{out}}}\right)^{1/d}
\end{equation}
where $d$ is the network depth, i.e. the number of layers excluding
the input layer. Thus, the network structure is fixed by $n_{\mathrm{out}}$
and $d$. \textcolor{black}{We tested different values for $d$ ranging
from 2 to 11; For alanine dipeptide, Supplementary Fig. 2b reports
the results in terms of the training success rate described in the
results section.} Networks have a number of parameters that ranges
between 100 and 400000, most of which are between the first and second
layer due to the rapid dimension reduction of the network. To avoid
overfitting, we use dropout during training \cite{JMLR:v15:dropout},
and select hyper-parameters using the VAMP-2 validation score. 

\paragraph*{Neural network hyperparameters}

Hyper-parameters include the regularization factors for the weights
of the fully connected and the Softmax layer, the dropout probabilities
for each layer, the batch-size, and the learning rate for the Adam
algorithm. Since a grid search in the joint parameter space would
have been too computationally expensive, each hyper-parameter was
optimized using the VAMP-2 validation score while keeping the other
hyper-parameters constant. We started with the regularization factors
due to their large effect on the training performance, and observed
optimal performance for a factor of $10^{-7}$ for the fully connected
hidden layers and $10^{-8}$ for the output layer;\textcolor{blue}{{}
}\textcolor{black}{regularization factors higher than $10^{-4}$ frequently
led to training failure. Subsequently, we tested the dropout probabilities
with values ranging from 0 to $\unit[50]{\%}$ and found $\unit[10]{\%}$
dropout in the first two hidden layers and no dropout otherwise to
perform well. The results did not strongly depend on the training
batch size, however, more training iterations are necessary for large
batches, while small batches exhibit stronger fluctuations in the
training score. We found a batch-size of $4000$ to be a good compromise,
with tested values ranging between $100$ and $16000$. The optimal
learning rate strongly depends on the network topology (e.g. the number
of hidden layers and the number of output nodes). In order to adapt
the learning rate, we started from an arbitrary rate of 0.0}5. If
no improvement on the validation VAMP-2 score was observed over 10
training iterations, the learning rate was reduced by a factor of
$10$. This scheme led to better convergence of the training and validation
scores and better kinetic model validation compared to using a high
learning rate throughout.

The time lag between the input pairs of configurations was selected
depending on the number of output nodes of the network: larger lag
times are better at isolating the slowest processes, and thus are
more suitable with a small number of output nodes. The procedure of
choosing network structure and lag time is thus as follows: First,
the number of output nodes $n$ and the hidden layers are selected,
which determines the network structure as described above. Then, a
lag time is chosen in which the largest $n$ singular values (corresponding
to the $n-1$ slowest processes) can be trained consistently.

\paragraph*{VAMPnet training and validation}

\textcolor{black}{We pre-trained the network by minimizing the negative
VAMP-1 score during the first third of the total number of epochs,
and subsequently optimize the network with VAMP-2 optimization (Sec.
\ref{subsec:VAMPnet}). In order to ensure robustness of the results,
we performed 100 network optimization runs for each problem. In each
run, the dataset was shuffled and randomly split into 90\%/10\% for
training and validation, respectively. To exclude outliers, we then
discarded the best 5\% and the worst 5\% of results. Hyperparameter
optimization was done using the validation score averaged over the
remaining runs. Figures report training or validation mean and 95\%
confidence intervals.}

\paragraph*{Brownian dynamics simulations}

The asymmetric double well and the protein folding toy model are simulated
by over-damped Langevin dynamics in a potential energy function $U(\mathbf{x})$,
also known as Brownian dynamics, using an forward Euler integration
scheme. The position $\mathbf{x}_{t}$ is propagated by time step
$\Delta t$ via:
\begin{equation}
\mathbf{x}_{t+\Delta t}=\mathbf{x}_{t}-\Delta t\frac{\nabla U(\mathbf{x})}{kT}+\sqrt{2\Delta tD}\mathbf{w}_{t},\label{eq:Brown}
\end{equation}
where $D$ is the diffusion constant and $kT$ is the Boltzmann constant
and temperature. Here, dimensionless units are used and $D=1$, $kT=1$.
The elements of the random vector $\mathbf{w}_{t}$ are sampled from
a normal distribution with zero mean and unit variance.

\paragraph*{\textcolor{black}{Hardware used and training times}}

\textcolor{black}{VAMPnets were trained on a single NVIDIA GeForce
GTX 1080 GPU, requiring between 20 seconds (for the double well problem)
and 180 seconds for NTL9 for each run.}

\paragraph*{Code availability}

TICA, $k$-means and MSM analyses were conducted with PyEMMA version
2.4, freely available at \href{http://pyemma.org}{pyemma.org}. VAMPnets
are implemented using the freely available packages\textit{ keras
\cite{chollet2015keras} }with \textit{tensorflow-gpu \cite{tensorflow2015-whitepaper}
}as a backend. The code can be obtained at \url{https://github.com/markovmodel/deeptime}.

\paragraph*{Data availability}

Data for NTL9 can be requested from the autors of \cite{LindorffLarsenEtAl_Science11_AntonFolding}.
Data for all other examples is available at \url{https://github.com/markovmodel/deeptime}.

\vspace{0.5cm}

\paragraph*{Author contributions}

A.M. and L.P. conducted research and developed software. H.W and F.N
designed research and developed theory. All authors wrote the paper.

\paragraph*{Acknowledgements}

We are grateful to Cecilia Clementi, Robert T. McGibbon and Max Welling
for valuable discussions. This work was funded by Deutsche Forschungsgemeinschaft
(Transregio 186/A12, SFB 1114/A4, NO 825/4-1 as part of research group
2518) and European Research Commission (ERC StG 307494 ``pcCell'').
\end{document}